\documentclass[letterpaper]{article} 
\usepackage[]{aaai23}  
\usepackage{times}  
\usepackage{helvet}  
\usepackage{courier}  
\usepackage[hyphens]{url}  
\usepackage{graphicx} 
\usepackage{natbib}  
\usepackage{caption} 
\usepackage{algorithm}
\usepackage{algorithmic}

\usepackage{newfloat}
\usepackage{listings}

\usepackage{booktabs}
\usepackage{multirow}
\usepackage{caption}
\usepackage{cite}
\usepackage{pifont}
\newcommand{\cmark}{\ding{51}}%
\newcommand{\xmark}{\ding{55}}
\usepackage{hyperref}
\usepackage{textcomp}
\usepackage{amsmath}
\usepackage{amsfonts}
\usepackage[title]{appendix}

\DeclareCaptionStyle{ruled}{labelfont=normalfont,labelsep=colon,strut=off} 
\floatstyle{ruled}
\newfloat{listing}{tb}{lst}{}
\floatname{listing}{Listing}
\title{AudioEar: Single-View Ear Reconstruction for Personalized Spatial Audio}
\author{
    Xiaoyang Huang\textsuperscript{\rm 1}, 
    Yanjun Wang\textsuperscript{\rm 1},
    Yang Liu,
    Bingbing Ni\textsuperscript{\rm 1}\thanks{Corresponding Author: Teng Li, Bingbing Ni}, \\
    Wenjun Zhang\textsuperscript{\rm 1},
    Jinxian Liu\textsuperscript{\rm 1},
    Teng Li\textsuperscript{\rm 2}$^\ast$
}
\affiliations{
    \textsuperscript{\rm 1}Shanghai Jiao Tong University, Shanghai 200240, China,
    \textsuperscript{\rm 2}Anhui University

    \{huangxiaoyang, nibingbing\}@sjtu.edu.cn
}

\usepackage{bibentry}

\begin{document}

\maketitle

\begin{abstract}
Spatial audio, which focuses on immersive 3D sound rendering, is widely applied in the acoustic industry. One of the key problems of current spatial audio rendering methods is the lack of personalization based on different anatomies of individuals, which is essential to produce accurate sound source positions. In this work, we address this problem from an interdisciplinary perspective. The rendering of spatial audio is strongly correlated with the 3D shape of human bodies, particularly ears. To this end, we propose to achieve personalized spatial audio by reconstructing 3D human ears with single-view images. First, to benchmark the ear reconstruction task, we introduce AudioEar3D, a high-quality 3D ear dataset consisting of 112 point cloud ear scans with RGB images. To self-supervisedly train a reconstruction model, we further collect a 2D ear dataset composed of 2,000 images, each one with manual annotation of occlusion and 55 landmarks, named AudioEar2D. To our knowledge, both datasets have the largest scale and best quality of their kinds for public use. Further, we propose AudioEarM, a reconstruction method guided by a depth estimation network that is trained on synthetic data, with two loss functions tailored for ear data. Lastly, to fill the gap between the vision and acoustics community, we develop a pipeline to integrate the reconstructed ear mesh with an off-the-shelf 3D human body and simulate a personalized Head-Related Transfer Function (HRTF), which is the core of spatial audio rendering. Code and data are publicly available\footnote{\url{https://github.com/seanywang0408/AudioEar}}.
\end{abstract}

\section{Introduction}
Spatial audio is widely applied in virtual reality, gaming, and movie production (\citeauthor{begault20003}), for distinguishing sound source positions and generating immersive 3D sound. Without it, people would lose the spatial sense of sound. The rendering of spatial audio depends on the Head-Related Transfer Function (HRTF) (\citeauthor{elliott2018head,blauert1997spatial}). HRTF is adopted in mid-to-high-end audio equipment, such as stereos, headphones, Hi-Fi, and so on. HRTF varies from one person to another since it depends on the 3D structure of human bodies including ears, head, and torso. While a personalized HRTF could be approximated by acoustic simulation (\citeauthor{pierce2019acoustics}) given a complete 3D human body structure, it is laborious to model human bodies using 3D scanners. Single-view reconstruction is an alternative solution. It is well studied to reconstruct personalized human heads and torsos within single-view images. However, without the modeling of ear structure, which is the central organ in the human hearing system, the simulated HRTF is still biased. In this work, we propose to obtain personalized HRTF by reconstructing the 3D mesh (\citeauthor{huang2022representation}) of human ears within a single-view image. \footnote{Strictly speaking, the ear structure include outer ear, middle ear, and inner ear. Our work only focuses on the visible structure (pinna) in the outer ear. Yet we still use the more understandable name \textit{ear} in the paper, following prior works (\citeauthor{jin2013creating}).} 

To benchmark the task of ear reconstruction, we collect a high-quality 3D ear dataset with an advanced structured-light 3D scanner, named AudioEar3D. Compared to prior evaluation protocol that uses 2D ear landmark re-projection error as the metric, a 3D benchmark is more precise. AudioEar3D includes $112$ ear point cloud scans with RGB images from $56$ individuals, each scan with $100,000$ to $250,000$ points, which capture high-resolution shape characteristics of ears. To the best of our knowledge, this is the largest and most accurate 3D ear dataset that is publicly available, as analyzed in Section \ref{sec:prior_dataset}. We believe that except for spatial audio, this dataset could also contribute to other applications, including digital humans (\citeauthor{kappel2021high}), morphology (\citeauthor{krishan2019study}) and medical surgery research (\citeauthor{varman2021ability}). 


Prior reconstruction methods for other parts of human body train on large-scale 2D datasets self-supervisedly (\citeauthor{zhang2020image, chen2019learning}). However, existing 2D ear datasets either lack semantic annotations or suffer from a small scale. This situation greatly limits the extension of CV studies on human ears. To this end, we build a large-scale 2D ear images dataset, including $2,000$ high-resolution ear images. They are selected from the FFHQ human face dataset, each of which accompanies manual annotations of occlusion and $55$ ear landmarks. The landmarks annotations greatly enrich the semantics of the ear images, enabling their usage in extensive applications.

Based on the self-supervised reconstruction pipeline, we proposed AudioEarM, a depth-guided reconstruction method tailored for ears. Due to the lack of public ear texture, which hinders self-supervised training, we first build an ear texture space from DECA (\citeauthor{feng2021learning}) for the UHM Ear (\citeauthor{ploumpis2020towards}), the ear shape model we use in our work. We excavate UV coordinate correspondence between UHM Ear vertices and the texture of DECA via K-Nearest-Neighbors searching and distance-based weighting. Besides, motivated by the characteristics of ear data, we design two loss functions: (i) contour loss which better accommodates the annotation error than landmark loss; (ii) similarity loss which encourages the network to predict distinct shapes for different samples. Moreover, we introduce synthetic data to improve reconstruction quality. Instead of directly combining synthetic data with real data and using 3D supervision to train the model, we use the synthetic data to train a monocular depth model to extract depth feature from ear images. Then we inject the depth information into the main network to guide the whole reconstruction process.


Lastly, to fill the gap between ear reconstruction and acoustic applications, we develop a pipeline to integrate the reconstructed ear into a 3D mesh human torso for acoustic simulation. We propose \textit{approximated Delaunay triangulation} on the 3D vertices to ensure the simulation validity of the integrated body mesh. We obtain the personalized HRTF with the reconstructed ear via simulation, and further demonstrate the effectiveness of the reconstruction model over the baseline by comparison on HRTF.









\section{Related Work}
\subsection{Spatial Audio}

Spatial audio, or spatial hearing, is a long-standing subject in acoustics (\citeauthor{culling2012spatial, vorlander2020auralization}). It focuses on how humans localize the position of acoustic sources and how to reproduce the spatial sense with a sound system. Humans locate the sound sources by \textit{biaural} and \textit{monaural} cues. Biaural cues include interaural time difference (ITD) and interaural level difference (ILD), which means that the arriving sound signals are filtered by the diffraction and reflection of the human body, including ear (pinna), head, and torso, leading to a difference of time, and intensity of arrival. Monaural cue is the spectral distortion of sounds. Further, the 3D shape of the human body varies between individuals, resulting in different binaural and monaural cues between different people. To model different spatial hearing characteristics between people, the Head-Related Transfer Function (HRTF) (\citeauthor{li2020measurement, xie2015typical}) is proposed and widely used in spatial sound rendering. HRTF describes the Sound Pressure Level (SPL, dB) of a sound source in each direction for a specific individual.
It is used to convert an arbitrary sound to a specific position as if the sound is originated from there. However, obtaining an accurate personalized HRTF is laborious since the measurement requires expensive equipment and special acoustic laboratories. Current implementations tend to deploy an average HRTF or choose one from an HRTF database (\citeauthor{guo20203d}). Besides physical measurement, HRTF could also be numerically simulated (\citeauthor{conrad2018hats, jensen2009improving}) based on Boundary Element Method (BEM). \citeauthor{meshram2014p} proposed to simulate a personalized HRTF by reconstructing the 3D human body using multi-view stereo (MVS). However, the reconstructed body is heavily blurred due to the imperfection of the reconstruction method, yielding large HRTF error. 


\subsection{Ear Datasets in 3D and 2D}
\label{sec:prior_dataset}
\begin{table}[t]
    \centering
    \setlength\tabcolsep{1pt}
    \caption{AudioEar3D and its counterparts. Quality is assessed based on the precision of acquisition equipments. }
    \vspace{-8pt}
    \begin{tabular*}{\hsize}{@{}@{\extracolsep{\fill}}lcccc@{}}
    \toprule
    3D Ear Dataset & Scale & with Image & Quality & Accessibility \\
    \midrule
    UND-J2      & $1,800$   & \cmark & $\star$ & \cmark \\
    York3DEar & $500$   & \xmark & $\star$ & \cmark \\
    SYMARE-1  & $20$     & \xmark & $\star\star\star$ & \cmark\\
    SYMARE-2  & $102$ & \xmark & $\star\star\star$ & \xmark\\
    Ploumpis et al.  & $234$ & \xmark & $\star\star\star$ & \xmark\\
    \textbf{AudioEar3D} &  $112$ &\cmark & $\star\star\star~\star$ & \cmark \\
    \bottomrule
    \end{tabular*}

    \label{tab:audioear3d}
    
    \caption{Comparisons of AudioEar2D and its counterparts. \textit{Lmks.} denotes landmark annotation. \textit{Occ.} means annotations of whether the ear is partially occluded by hair or earrings.}
    \small
    \begin{tabular*}{\hsize}{@{}@{\extracolsep{\fill}}lccccc@{}}
    \toprule
    \multirow{2}{*}{2D Ear Dataset} & \multirow{2}{*}{Scale} & \multirow{2}{*}{Source} & \multicolumn{2}{c}{Annotations} & \multirow{2}{*}{Usage}\\
    & & & Lmks. & Occ. \\
    \midrule
    UND-E  & $464$ &  Limited & \xmark & \xmark & Biometrics\\
    AMI  & $700$& Limited  & \xmark & \xmark & Biometrics\\
    IIT Delhi Ear & $754 $ & Limited & \xmark& \xmark & Biometrics \\
    WPUTEDB  & $3,348$ & Limited & \xmark& \xmark & Biometrics \\
    UBEAR   & $4,410$ & Limited & \xmark & \xmark & Biometrics\\
    IBug-B  & $2,058$ & In-the-wild & \xmark& \xmark & Biometrics \\
    AWE  & $9,500$ & In-the-wild & \xmark& \xmark & Biometrics\\
    EarVN  & $28,412$ & In-the-wild & \xmark& \xmark & Biometrics \\
    IBug-A  & $605$ & In-the-wild & \cmark & \xmark & Reconstruction\\
    \textbf{AudioEar2D} & $2,000$ & In-the-wild & \cmark & \cmark & Reconstruction\\
    \bottomrule
    \end{tabular*}
    \label{tab:audioear2d}
\end{table}

Previous CV studies on human ears are mostly confined to biometric application (\citeauthor{o20203d}), and do not prevail as other parts of human body, such as faces, hands and skeletons. This leads to the lack of a rich ear dataset in 3D and 2D, as stated in Table \ref{tab:audioear3d}, \ref{tab:audioear2d}. 

Existing 3D ear datasets suffer from either non-accessibility, small scale, or low quality. \citeauthor{yan2007biometric} collect $1,800$ ear depth maps in a resolution of $640 \times 480$ using a depth sensor. Since the depth maps are single-view, they do not represent complete ear shapes. \citeauthor{dai2018data} publish York3DEar, which is composed of $500$ deformed 3D ear meshes. These meshes are not collected with instruments, but are estimated by a data-augmenting technique based on a 2D ear dataset, which introduces unforeseeable error. \citeauthor{jin2013creating} propose the SYMARE ear database consisting of the measured HRIR and the upper torso, head and ear mesh model collected by magnetic resonance imaging (MRI) from $61$ individuals. However, only $10$ (SYMARE-1) individuals' data are accessible, while the rest (SYMARE-2) are kept private. \citeauthor{ploumpis2020towards} collect $121$ ear models from $64$ adults and $133$ ears models from children via CT scans. However, these data are not publicly available. Besides, most of the above 3D ear datasets lack the corresponding ear images (except for UND-J2), making it unfeasible to perform single-view reconstruction tasks upon them. Moreover, most of these data are collected by MRI, whose measurement error is generally about $1$ millimeter (\citeauthor{anna2018the}). This measurement error is not negligible for such an elaborately-structured organ, thereby introducing additional error in the simulation of HRTF. 

Existing 2D ear datasets include 
UND-E (\citeauthor{chang2003comparison}), 
AMI (\citeauthor{esther}), 
IIT Delhi Ear (\citeauthor{kumar2012automated}), 
WPUTEDB  (\citeauthor{frejlichowski2010west}), 
UBEAR (\citeauthor{raposo2011ubear}), 
IBug(-A/B) (\citeauthor{zhou2017deformable}), 
AWE (\citeauthor{emervsivc2017unconstrained}) and
EarVN (\citeauthor{hoang2019earvn1}), 
most of which are for human identity recognition. The summarization of these datasets is shown in Table \ref{tab:audioear2d}. 


\subsection{3D Morphable Models Reconstruction}
A 3D Morphable Model (3DMM) is a parametric model that encodes the shape and texture of 3D meshes into a latent space. The most studied structure in 3DMM reconstruction are human faces (\citeauthor{ploumpis2020towards, ploumpis2019combining}), hands (\citeauthor{chen2021model, wang2020rgb2hands, zhang2020mediapipe}) and bodies (\citeauthor{choi2020pose2mesh, corona2021smplicit}). To achieve plausible performance, the 3DMM reconstruction algorithms require a large set of 2D images for self-supervised training, usually accompanied by semantic annotations, such as landmarks or poses. They use a feature extractor and a multi-layer perceptron (MLP) to regress the latent code of shape and texture. The latent codes are fed into the differentiable 3DMM models to obtain colored 3D meshes. Then the meshes are projected to images with a differentiable renderer (\citeauthor{kato2018neural, huang2022boosting}). The photometric loss on images and the landmark re-projection loss are minimized to train the network. The popular face reconstruction algorithm DECA (\citeauthor{feng2021learning}) proposes to conduct robust detail reconstruction by regressing a UV displacement map. \citeauthor{sun2020human} propose to reconstruct 3D ear meshes with the YEM model (\citeauthor{dai2018data}). They derived a color model from images and minimized the photometric and landmark error to regress shape latent codes.

\section{Method}
\subsection{Data collection}
We collect two ear datasets for personalized spatial audio. One is AudioEar3D, a 3D ear scan dataset, for benchmarking the ear reconstruction task. The other is AudioEar2D, for the training of ear reconstruction models. 
\subsubsection{AudioEar3D} Among various 3D scanning equipments, we choose a structured-light 3D scanner to collect our 3D ear data, for the following reasons: (i) Compared to previous 3D ear scanning methods, such as MRI and CT, a structured-light scanner is much more convenient and accessible (it can be integrated into a small portable device), which makes it easy to cover a larger population; (ii) The operation and post-processing are simpler; (iii) The resolution of structured-light scanners is higher compared to others, under the same price. Specifically, we use MantisVision{\textregistered} F6-SR \footnote{Product introduction: \url{https://mantis-vision.com/handheld-3d-scanners/}}, a portable high-resolution 3d scanner, which is designed to be able to scan detailed organ models in medical applications. Its plane resolution and depth resolution are $0.1$ mm and $0.4$ mm respectively (compared to a resolution of $1$ mm in most MRI scanners), with a frame rate of 8 FPS.

\begin{figure*}[!t]
    \centering
    \includegraphics[width=0.93\linewidth]{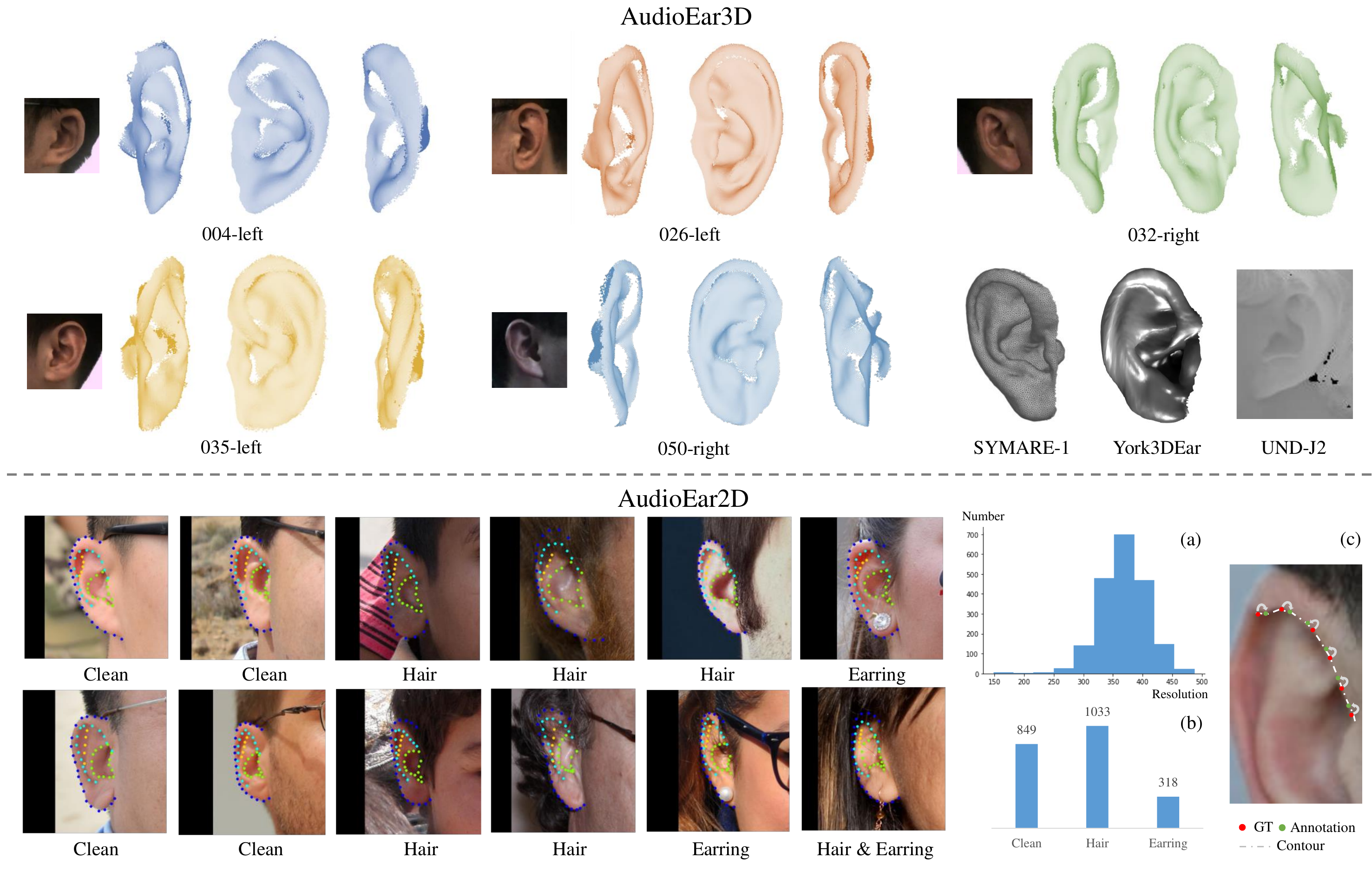}
    \caption{Visualization of the two collected datasets. \textbf{AudioEar3D} (top): We show five samples with their three-view rendered point clouds and RGB images. Compared to existing 3D ear dataset, AudioEar3D is high-quality, large-scale, and also publicly available. \textbf{AudioEar2D} (bottom): We illustrated several samples with their occlusion annotations and the 55 landmarks. The four colors of the landmarks indicate four main contours of the ears. We flip the left ears to the right for better visualization. }
    \label{fig:dataset}
\end{figure*}

To alleviate the self-occlusion issue brought by the complex ear structure and obtain a clear and complete 3D ear scan, we scan each individual with the above device for about $90$ seconds from different directions. We remove the other parts of the body and only keep the ear part. This yields approximately $100,000$ to $250,000$ points for each ear scan, which captures high-resolution shape characteristics of ears. Besides, we take frontal pictures of the ears from a distance similar to the scanner. At present we have obtained data of $112$ ears from $56$ subjects. We plan to further extend the scale to cover a larger population. We show several samples in Figure \ref{fig:dataset}. The out-coming dataset is completely \textit{anonymized} for privacy issues.


\subsubsection{AudioEar2D} 

We obtain our 2D ear dataset from a high-quality public human face dataset, Flickr-Faces-HQ Dataset (FFHQ) (\citeauthor{karras2019style}), which consists of $70,000$ face images in the resolution of $1,024\times 1,024$ that are various in terms of pose, age, and ethnicity. To find images that contain clear ears, we first train an ear detection model on the IBug Collection-B dataset (\citeauthor{zhou2017deformable}), which contains the bounding box of ears, from a pre-trained Yolov4 (\citeauthor{bochkovskiy2020yolov4}). Then we adopt the trained detection model on the FFHQ dataset to roughly sift out high-quality ear images according to detection confidence and bounding box area. Besides, we leverage WHENet (\citeauthor{zhou2020whenet}), a pre-trained head pose estimation model, to further remove those images that have bad view angles. We cut the remaining images into squares based on the bounding box. We preserve the original resolution of the cut ear image to avoid information loss induced by interpolation resizing. Thanks to the bounding box area filtering, the image resolution is mostly above $300\times 300$, as illustrated in Figure \ref{fig:dataset} bottom (a). Next, we manually filter the remaining images and annotate the $55$ landmarks, following a landmark protocol in IBug Ear (\citeauthor{zhou2017deformable}). Moreover, we annotate whether the ear is clean or is occluded by hair or by earrings, which could act as noise in some applications. The distribution of occlusion is shown in Figure \ref{fig:dataset} bottom (b). We obtain $2,000$ high-resolution ear images, each with $55$ annotated landmarks in the end.



\subsection{AudioEarM: Depth-Guided Ear Reconstruction}
We develop our method upon a common pipeline of parametric shape reconstruction, which regresses the latent code of the 3D shape, texture, camera, and lighting via self-supervised training. We first adapt this pipeline to ears with several effective modifications, which are motivated by the specific characteristic of ears. Besides, to further boost the performance, we propose to leverage synthetic ear images to train a depth estimation model and fuse the depth information into the reconstruction model via multi-scale feature alignment. We name our method AudioEarM.

\subsubsection{Texture Acquisition} 
We use the UHM Ear (\citeauthor{ploumpis2020towards}), the best-quality ear 3DMM to transform the shape latent code into a 3D mesh. The UHM Ear reduces the N vertices ($2800$) of an ear mesh to a lower dimension ($236$) using PCA. Let $\overline{\mathbf{S}} \in \mathbb{R}^{3N}$ denotes the mean shape, $\vec{v} \in \mathbb{R}^{236}$ and $\mathcal{U} \in \mathbb{R}^{3N \times236}$ denote the eigenvalues and eigenvectors of the 3DMM. The shape variations are modeled by eigenvalue-weighted linear blendshapes: $B_{\mathcal{U}}(\vec{\beta} ; \mathcal{U})=\sum_{n=1}^{236} v_{n} \beta_{n} \mathcal{U}_{n}$, where $\vec{\beta} \in \mathbb{R}^{236}$ is the predicted shape latent code. The final ear mesh is obtained by:
\begin{equation}
M(\vec{\beta})=\sum_{n=1}^{236} v_{n} \beta_{n} \mathcal{U}_{n} + \overline{\mathbf{S}} 
\end{equation}
The UHM Ear does not provide texture, which is indispensable to minimize the photometric loss in a self-supervised reconstruction pipeline. We extract the ear texture space from DECA (\citeauthor{feng2021learning}) to enable colored rendering. In DECA, given a texture latent code $\vec{\theta} \in \mathbb{R}^{|\vec{\theta}|}$, a texture map $T(\boldsymbol{\theta}) \in \mathbb{R}^{h \times w \times 3}$ ($h\times w$ is the resolution) could be calculated. Each vertex in the DECA mesh corresponds to a UV coordinate ${p}_{deca}$ in the texture map. However, the vertices of the UHM ear do not have correspondence with DECA. To this end, we first visually align the mean UHM ear with the ear in DECA as close as possible. For each vertex $v_{uhm} \in \mathbb{R}^{3}$ in UHM model, we then search its K-Nearest-Neighbors $v_{deca}^i\in \mathbb{R}^{3}, i=1,2\dots,k$ in DECA vertices (k is $3$ in our implementation). We assign the KNN-distance-weighted UV coordinates to the UHM vertices based on Euclidean distance between $v_{uhm}$ and $v_{deca}^i$. Formally, the UV coordinates of $v_{uhm}$ are computed by:
\begin{equation}
{p}_{uhm} = \sum_{i=1}^{k} \frac{D(v_{uhm},v_{deca}^i)}{\sum_{j=1}^{k} D(v_{uhm},v_{deca}^j)} {p}_{deca}^i
\end{equation} 

\paragraph{Contour Loss.} Contour loss is concerned with the four contours formed by the 2D landmarks. (The four contours are indicated by four different landmark colors in AudioEar2D in Figure \ref{fig:dataset}.) As illustrated in Figure \ref{fig:dataset} bottom (c), the landmark annotation have bias \textit{along} the contour, while the contour is well represented by the annotations. The reason is that the contour is more visually salient for human eyes. Hence measuring the error of contour prediction in training is superior to landmark error. Practically, we connect the landmarks one by one to form four polylines and uniformly re-sample dense points on these polylines. We measure the chamfer distance of the re-sampled points $\mathcal{P}$ between the ground truth and prediction as contour loss: 
\begin{equation}
    L_{contour} = \frac{1}{4}\sum_{i=1}^4ChamferDistance(\mathcal{P}_{gt}^i, \mathcal{P}_{pred}^i)
\end{equation}
\noindent where the superscript $i$ is the index of the four contours.


\begin{figure*}[!ht]
    \centering
    \includegraphics[width=\linewidth]{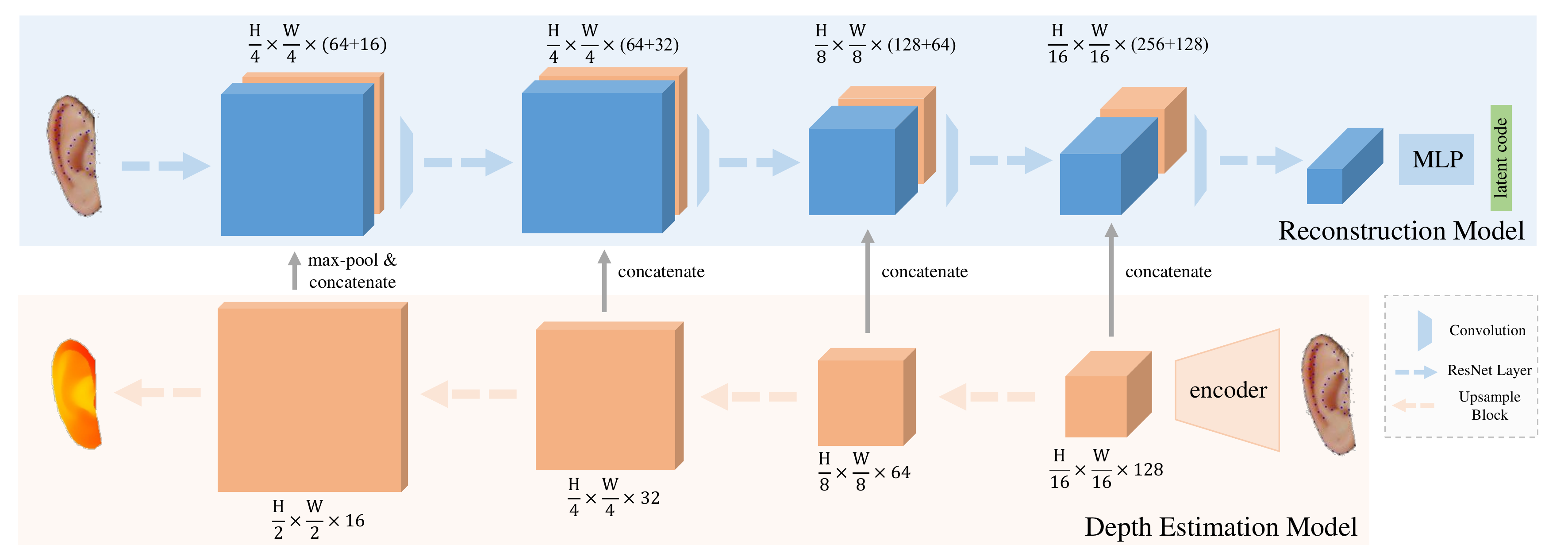}
    \caption{The architecture of our depth-guided reconstruction model, AudioEarM. A depth estimation model, which is trained on a synthetic dataset, guides the main reconstruction network with multi-scale feature alignment.}
    \label{fig:audioearm}
    \vspace{-8pt}
\end{figure*}

\paragraph{Similarity Loss.} In our early experiments, we find that the predicted shape latent codes are too similar across all instances. This similarity is undesirable since the shape should be distinct across different individuals. We encourage the network to predict distinct shapes by measuring the mean cosine similarity between the shape latent codes in a batch and penalizing high similarity: 
\begin{equation}
L_{sim}=\frac{1}{bs \times(bs-1)} \sum_{i, j=0, i \neq j}^{bs-1} \frac{\beta_i \beta_j}{|\beta_i||\beta_j|}
\end{equation}
\noindent where $bs$ denotes the batch size and $\beta_i, \beta_j$ denote the $i$th, $j$th sample in a batch.





Besides, we prevent extreme rotation angles and scale parameters in camera with $L1$-penalization, since the angle and scale of ears in images are generally within a certain scope. We combine these three loss functions with the commonly used losses: landmark, photometric loss, mesh smooth loss and latent code regularization. The overall loss is the weighted sum of the losses above. See appendix for a complete formulation of all loss functions.


\subsubsection{Depth-Guided Reconstruction} \citeauthor{zhang2020mediapipe} demonstrates that introducing synthetic data with 3D supervision into trainset could improve the quality of single-view hand reconstruction. Yet we empirically find that directly migrating this approach to our task brings negative effect (see ablation study). Alternatively, we leverage the power of pre-training (\citeauthor{yang2021reinventing}), that is to use synthetic data to pre-train a monocular depth estimation (MDE) model. Then we leverage the depth feature extracted by the MDE model to facilitate the whole reconstruction process, as shown in Figure \ref{fig:audioearm}. 

To generate a synthetic ear dataset with 3D supervision, we randomly sample shape and texture latent codes from a Gaussian distribution, and render corresponding RGB images and depth maps. We generate $10,000$ samples in this way. Then we train an MDE model on this synthetic dataset, using the depth map as direct supervision. The MDE model (\citeauthor{laina2016deeper}) consists of an encoder and a decoder (Figure \ref{fig:audioearm}). The ResNet-34 (\citeauthor{he2016deep}) encoder extracts high-level feature from images, while the decoder predicts a depth map from the feature. The extracted feature is upsampled by the decoder layer by layer until the original resolution is restored. We aim to leverage the depth information obtained by decoder to guide the reconstruction process. To this end, we concatenate the multi-scale upsampled feature with the feature before each layer in the main network. We insert an additional convolution layer each time after concatenation to transform the increased channel numbers into original ones, so as to match the channel of the next layer. Since the last-layer resolution of the MDE model is twice as big as the first-layer resolution in the main network, we conduct max-pooling on the last-layer MDE feature before concatenation. In this way, we explicitly fuse the depth information into the main network to guide the reconstruction process. Ablation study shows that this method brings notable improvement.

\begin{figure*}[!ht]
    \centering
    \includegraphics[width=\linewidth]{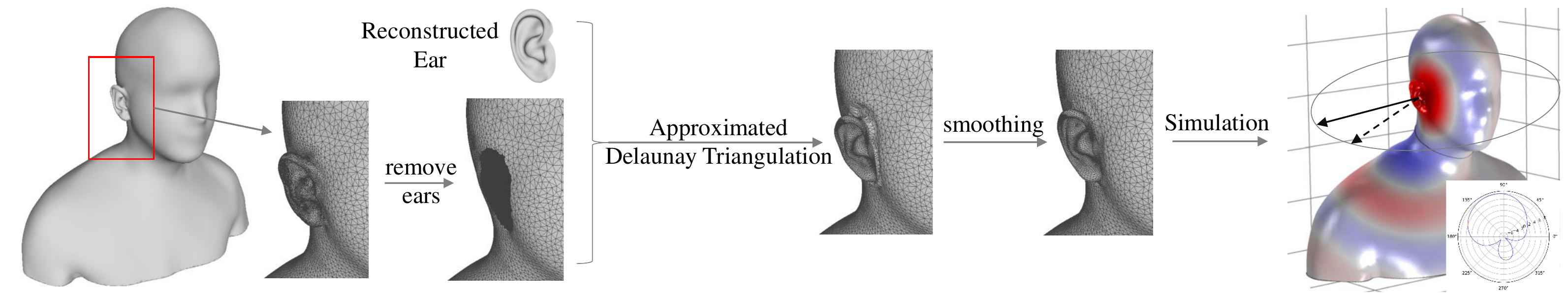}
    \caption{The proposed integration and simulation pipeline. We conduct \textit{approximated Delaunay triangulation} to combine the reconstructed ear with the body and simulate the personalized HRTF in all azimuth angles ($361$ angles) on the horizontal plane.}
    \label{fig:simulation_pipeline}
\end{figure*}

\subsection{Acoustic Simulation}

Spatial audio rendering depends on the HRTF, which is defined as the Sound Pressure Level (SPL) measured at the eardrum (or the ear canal entrance in practice). We simulate a personalized HRTF with COMSOL\textsuperscript{TM}, a Multiphysics software that uses the Boundary Element Method (BEM) for simulation. However, the simulation requires a complete human upper body, not just an ear. To this end, we develop a pipeline to integrate our reconstructed ears into an off-the-shelf human body 3D model (\citeauthor{braren2020high, braren2020measurement}). The pipeline is illustrated in Figure \ref{fig:simulation_pipeline}.

We manually remove the original ears of the given 3D body and place the reconstructed ear to the consequent hole. To combine the ear and body mesh together, we propose a method named \textit{approximated Delaunay triangulation}. Delaunay triangulation is a triangulation method for 2D point sets, which maximize the minimal angle of all the angles of the triangles in the triangulation process, such that sliver triangles are avoided. Sliver triangles are inappropriate for BEM simulation. Since both the edges of the ear and body hole are not on a 2D plane (but close to a plane), we project the edges to a plane which is found by the averaged normals of the mesh faces around the edges. In this way, we can conduct \textit{approximated Delaunay triangulation} on the vertices of the two edges and fix the crack between ears and heads. Since the geometry of the edges is not greatly changed by the projections, the sliver triangles could still be avoided as much as possible. We leverage the Detri2 program \footnote{\url{http://www.wias-berlin.de/people/si/detri2.html}} for Delaunay triangulation. At last, we smooth the generated mesh and use it for simulation. We simulate the HRTF in all azimuth angles ($361$ angles in total) on the horizontal plane for three frequencies (f = 1033.6 Hz, 2067.5 Hz, and 3962.1 Hz), following a general protocol. Note that the simulation pipeline involves manual efforts and is not part of the evaluation protocol of ear reconstruction, but for application use. We evaluate reconstruction performance on AudioEar3D.


\section{Experiments}


\subsection{Ear Reconstruction Benchmark}
\label{sec:recon}
\subsubsection{Evaluation} We leverage the full AudioEar3D dataset to evaluate the performance of ear reconstruction models. We compute the distance from each point in the ground-truth ear scan to the predicted ear mesh. (The distance from a point to a mesh is defined as the distance from the point to the closest triangular face in the mesh.) We average the distance across all points in the scan as the evaluation metric, named scan2mesh (S2M). However, since the scale and position are not aligned between the scans and the predicted meshes, we first register the mesh with the scan in a gradient-based method. We iteratively optimize scale and position parameters for registration. To reduce the evaluation time and avoid convergence in a local minimum, we design a three-stage registration scheme. First, we minimize four manually chosen key points to obtain a coarse registration. Next, we randomly sample points on the predicted mesh surfaces and minimize the chamfer distance between the sampled points and the scans. Last, we directly minimize the S2M distance between the scan and mesh. Since the dense points clouds in AudioEar3D would cause intense computation, we randomly down-sample the scan to $1,000$ points for the S2M and only compute the S2M distance on the whole scan in the last iteration. We use Adam (\citeauthor{kingma2014adam}) with an initial learning rate of $0.45$ for optimization. Each stage lasts for $166$ iterations. The learning rate is multiplied by $0.1$ when entering the next stage. After numerous tests, we empirically find that this setting can make most registrations converge to a global minimum. We conduct the registration process on all the $112$ samples in AudioEar3D, which takes about half an hour, and average the S2M across all samples as the final evaluation metric. The left ear scans are transformed to the right ones by reflection in the sagittal plane.


\subsubsection{Setting}
To compare, we send the average ear of UHM into the registration process, as a baseline without any personalization. We also compare our method with a prior ear reconstruction method named HERA (\citeauthor{sun2020human}). Since HERA did not publish their code and ear model, we implement it using our ear model. Besides, we would like to compare AudioEarM with the SOTA face reconstruction algorithm, DECA (\citeauthor{feng2021learning}). Since a detailed UV displacement map in DECA is not available for ears, we only implement the coarse branch in DECA, denoted as DECA-coarse. We describe the training configurations in the appendix.
\subsubsection{Results}

\begin{figure}[!t]
    \centering
    \includegraphics[width=\linewidth]{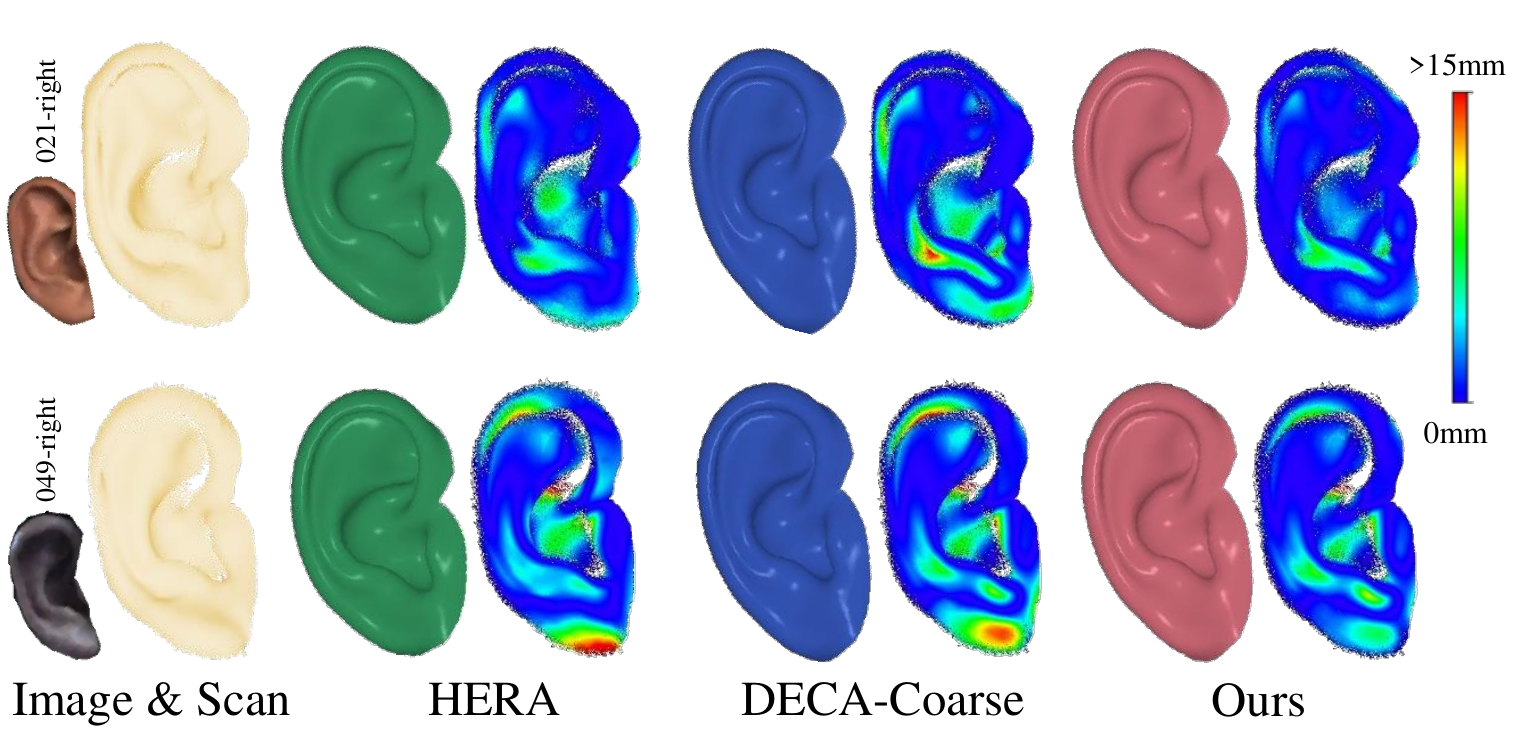}
    \caption{We visualize the \textbf{S2M error map} of our method and two baselines on testset (AudioEar3D). Our method has lower error compared to baselines.}
    \label{fig:results_vis}
\end{figure}

As listed in Table \ref{tab:ablation_model} \#6-9, our method surpasses the average ear baseline, HERA and DECA-coarse, with a final S2M distance of $1.28$ mm. The average ear yields an S2M of $1.78$ mm, while HERA and DECA-coarse yields an S2M of $1.70$ and $1.46$ mm. We visualize the \textbf{S2M error map} of our method and two baselines on testset (AudioEar3D). Our method has lower error compared to baselines.

\begin{table}[t]
    \centering
    \caption{Comparisons of reconstruction methods. \{\#1 + texture\} means texture is added on top of Method\#1. Similar meanings for other $+$,$-$ notations.}
    \vspace{-7 pt}
    \setlength\tabcolsep{10pt}
    \begin{tabular}{c|lc}
    \toprule
    \# & Model & S2M $\downarrow$\\
    \midrule
    
    1 & Naive & 1.73mm\\
    2 & \#1 $+$ texture & 1.65mm \\
    3 & \#2 $+$ similarity loss & 1.63mm \\
    4 & \#2 $+$ contour loss & 1.52mm \\
    5 & \#2 $+$ contour \& similarity & 1.47mm \\
    \textbf{6} & \textbf{Ours} (\#5 $+$ depth-guided)  & \textbf{1.28mm}\\
    7 & Avg ear & 1.78mm \\
    8 & HERA  & 1.70mm   \\
    9 & DECA-coarse  & 1.46mm   \\

    \bottomrule
    
    \end{tabular}
    \label{tab:ablation_model}
\end{table}

\subsection{Comparisons of Reconstruction Models}
We examine the impact of each part in our method (Table \ref{tab:ablation_model} \#1-6). All the models are trained with the same setting described in Section \ref{sec:recon}. We first implement a baseline model (Method\#1) in which a common self-supervised reconstruction pipeline is adopted to our data without modifications. A single skin color is assigned to the texture instead of our texture space. Method\#1 yields an S2M of $1.73$ mm, which has little improvement over the average ear. Then we replace the single color with our texture space (Method\#2), which reduces the S2M error to $1.65$ mm. On top of it, we add the similarity loss and contour loss into the training process (Method\#3-5). Adding the two losses together yields a result of $1.47$ mm. The effect of contour loss is more notable than the similarity loss. Finally, we use the depth estimation model to perform depth-guide reconstruction, and obtain an S2M of $1.28$ mm, which is our best result. The reason behind the superiority is that predicting a regular depth map is less ill-posed than predicting a shape latent code. Besides, depth could inform the object's geometry. Similar motivation is enlightening in pseudo-lidar approaches for 3D object detection (\citeauthor{qian2020end}).

\begin{table}[!t]
	\centering
    \caption{Results of HRTF simulation measured in SPL error. The first and second row in each sample denote \textit{Avg} and \textit{Pred} results respectively. We see that better reconstruction results bring more realistic HRTF. The predicted ears in the first 5 samples have lower S2M and also lower SPL error than average,  vice versa andfor the last 2 samples.}
    \vspace{-7pt}
    \setlength\tabcolsep{5pt}
    \small
    \begin{tabular}{l|c|ccc}
        
        \toprule
        \multirow{2}{*}{Sample} & S2M $\downarrow$ & \multicolumn{3}{c}{SPL Error $\downarrow$ (dB$\times$ 10)} \\
        & (mm) & f=1kHz & f=2kHz & f=4kHz \\
        \midrule
        \multirow{2}{*}{033-right} & 1.49 & 1.12$\pm$0.75 &                                          3.59$\pm$5.97 & 4.13$\pm$5.44\\
                                   & 0.94 & 0.33$\pm$0.19 & 0.59$\pm$0.89 & 2.32$\pm$2.84 \\
        \midrule
        \multirow{2}{*}{021-left} & 1.66 & 0.58$\pm$0.27 &                                        1.20$\pm$2.19 & 4.25$\pm$3.43\\
                                & 0.88 & 0.30$\pm$0.12 & 1.30$\pm$1.47 & 2.39$\pm$1.85 \\
        \midrule
        \multirow{2}{*}{036-left}& 2.23 & 1.23$\pm$1.17 &                                               3.52$\pm$9.16 & 6.97$\pm$11.97\\
                                  & 1.85 & 1.13$\pm$0.89 & 2.69$\pm$4.58 & 4.75$\pm$5.89 \\
        \midrule
        \multirow{2}{*}{008-left} & 1.23 & 1.69$\pm$1.31 &                                          4.50$\pm$6.71 & 5.29$\pm$5.40 \\
                                & 0.74 & 1.35$\pm$0.69 &   3.532$\pm$5.97 & 6.75$\pm$6.49 \\
        \midrule
        \multirow{2}{*}{048-left}  & 1.35 & 0.40$\pm$0.30 &                                          1.27$\pm$1.64 & 5.00$\pm$4.76 \\
                                 & 0.94 & 0.23$\pm$0.14 &   1.18$\pm$1.23 & 3.31$\pm$2.88 \\
        \midrule
        \multirow{2}{*}{040-right} & 1.27 & 1.00$\pm$0.66 &                                         2.83$\pm$5.65 & 5.54$\pm$10.23\\
                                & 1.58 & 1.17$\pm$0.81 &   3.57$\pm$6.54 & 5.99$\pm$10.43 \\
        \midrule
        \multirow{2}{*}{029-left}  & 1.33 & 2.11$\pm$0.98 &                                          5.95$\pm$9.76 & 8.06$\pm$9.47 \\
                                & 1.46 & 1.12$\pm$0.76 &   3.44$\pm$5.98 & 10.09$\pm$10.74 \\
        \bottomrule
    \end{tabular}
    \label{tab:simulation}
\end{table}

\section{HRTF Simulation}
\subsubsection{Evaluation.} In this part, we conduct the HRTF simulation using our predicted ears. We compare the HRTF of the predicted ears against the ground-truth ears. The HRTF of the ground-truth ears are simulated using the ear meshes registered from the raw ear scans in AudioEar3D. Specifically, we send the mean UHM Ear into the same registration process described in Section \ref{sec:recon}, except that the optimized parameters in the second and third stages include the shape latent code of UHM besides the original ones. The registration yields an average S2M of $0.11$ mm among all scans, which is fairly low. The low S2M error indicates the optimized meshes are rather close to the true ear shape. We measure the mean SPL error in absolute value across all angles between \textit{ground-truth} and predicted ones. As comparisons, we replace the predicted ears with the mean UHM Ear (denoted as \textit{Avg} in Table \ref{tab:simulation}). The SPL error of the mean UHM Ear represents the spatial audio effect without personalization. Larger errors indicate worse spatial audio. Since the simulation experiment is labour-intense and time-consuming, we only show the results of several samples for demonstration.

\subsubsection{Results.} Table \ref{tab:simulation} compares the results between \textit{Avg} and \textit{Pred} of each sample. The first and second row in each sample denote \textit{Avg} and \textit{Pred} results respectively. We see that better reconstruction results (lower S2M) bring more realistic HRTF (lower SPL error). For those samples whose reconstructed S2M distances are lower than the mean UHM Ear (the first $5$ samples), both the mean value and variation of SPL error are also lower, meaning that they obtain a closer HRTF to the ground truth than the mean UHM Ear. For the last two samples, the S2M of predicted ears is higher and the SPL error is somewhat larger, indicating that the obtained HRTF is highly related to the reconstruction results. It is necessary to develop advanced ear reconstruction algorithms for more realistic spatial sound effects in the coming VR trend. For a better understanding on the SPL error, we illustrate it with polar coordinates in the appendix. 

\section{Conclusion}

This work considers 3D ear reconstruction from single-view images for personalized spatial audio. For this purpose, we collect a 3D ear dataset for benchmarking and a 2D dataset for the training of ear reconstruction models. Besides, we propose a reconstruction method guided by a depth estimation network that is trained on synthetic data, with two loss functions tailored for ear data. Lastly, we develop a pipeline to integrate the reconstructed ear mesh with a human body for acoustic simulation to obtain personalized spatial audio. 

\section{Acknowledgement}

This work was supported by National Science Foundation of China (U20B2072, 61976137). This work was also partially supported by Grant YG2021ZD18 from Shanghai Jiao Tong University Medical Engineering Cross Research.

\bibliography{aaai23}




\clearpage

\begin{appendices}
\section{Appendix A: Details of Loss Functions}

\subsubsection{Contour Loss} We split the $55$ landmarks $K_i \in \mathbb{R}^{2},i=1,2\dots,55$ into $4$ groups according to their distribution on the ears. The groups are indicated in Figure 2 (AudioEar2D) by four different landmark colors. Then we connect the landmarks within one group one by one and get $4$ polylines. Then we uniformly sample $N$ points on each polylines, denoted as $\mathcal{P}^1, \mathcal{P}^2, \mathcal{P}^3, \mathcal{P}^4 \in \mathbb{R}^{N \times 2}$. We execute these steps on both the annotated landmarks and the projected landmarks from the predicted meshes, and obtain four pairs of sampled points $(\mathcal{P}_{proj}^1, \mathcal{P}_{gt}^1), (\mathcal{P}_{proj}^2, \mathcal{P}_{gt}^2), (\mathcal{P}_{proj}^3, \mathcal{P}_{gt}^3), (\mathcal{P}_{proj}^4, \mathcal{P}_{gt}^4)$. The contour loss is the average of the chamfer distance between $\mathcal{P}_{proj}$ and $\mathcal{P}_{gt}$ in each pair:

\begin{equation}
    L_{contour} = \frac{1}{4}\sum_{i=1}^4ChamferDistance(\mathcal{P}_{proj}^i, \mathcal{P}_{gt}^i)
\end{equation}

\noindent where the superscript $i$ is the index of the four contours.

\subsubsection{Camera Loss} We first define the range loss, which penalize a variable with $L2$-regularization when it is out of a specified range $[x_{min}, x_{max}]$:

\begin{equation}
\begin{split}
L_{range}(x,x_{min},x_{max}) = 
\begin{cases}(x_{min}-x)^{2} & \text { if } x<x_{min} \\ (x-x_{max})^{2} & \text { if } x>x_{max} \\ 0 & \text { otherwise }\end{cases}
\end{split}
\end{equation}

Based on the range loss, we constrain the scale $s$ and the rotation angle along y axis, $r_{y}$. The camera loss is formulated as:
$$
L_{cam}=L_{range}(s,0.5,4)+L_{range}(r_{y},-1.5,-0.5)
$$

\subsubsection{Landmark Loss}  We use the normalized landmark loss proposed by \citeauthor{zhou2017deformable}. The landmarks on the predicted 3D ear mesh $K_{pred}^i \in \mathbb{R}^{3}, i=1,2\dots,55$ are projected to the image plane by an orthogonal projection function $\mathbf{C}$. The landmark loss is calculated between the projected landmarks and the ground-truth 2D landmarks $K_{gt}^i \in \mathbb{R}^{2}, i=1,2\dots,55$ on the image plane. The landmark loss is formulated as:
\begin{equation}
    L_{lmk}=\frac{\sum_{i=1}^{55}\left\|K_{gt}^i-\mathbf{C}(K_{pred}^i)\right\|_{2}}{55\times D_{gt}}
\end{equation}
\noindent where $D_{gt}$ is the diagonal length of the ground-truth landmarks' bounding box.

\subsubsection{Photometric Loss} The photometric loss computes the pixel error between the input image $I$ and the rendered image $I_r$. To ignore the background of the input image, we use the 2D landmarks to form an ear mask $M_I$. Specifically, we connect the outer landmarks to form a closed polygon and fill the inner part of the polygon with value $1$, the outer part with value $0$. The loss is computed as 
\begin{equation}
    L_{p}= \Vert M_I\cdot I - I_r\Vert_2
\end{equation}

\subsubsection{Mesh Smooth Loss} We add laplacian smoothing (\citeauthor{desbrun1999implicit, nealen2006laplacian}) on the predicted ear meshes to avoid sharp deformation. We adopt uniform laplacian for simplicity. For each vertex $v_i$, we denote its neighbor vertices as $\mathcal{N}_{v_i}$ and the number of neighbor vertices as $N=|\mathcal{N}_{v_i}|$. The laplacian of $v_i$ is defined as:
\begin{equation}
\mathbf{Laplacian}(v_i) = \sum_{j=1}^{N} w_{i j}\left(\mathcal{N}_{v_i}^j-v_{i}\right)
\label{equ:lap1}
\end{equation}
For uniform laplacian, $w_{ij} = 1/|\mathcal{N}_{v_i}| = 1/N$. Therefore Equation \ref{equ:lap1} could be simplified as:
\begin{equation}
\mathbf{Laplacian}(v_i) = \frac{1}{N}\sum_{j=1}^{N}\mathcal{N}_{v_i}^j-v_{i}
\label{equ:lap2}
\end{equation}
The term $\frac{1}{N}\sum_{j=1}^{N}\mathcal{N}_{v_i}^j$ indicates the centroid of the neighbor vertices. The final mesh smoothing loss is the average on the laplacian of all vertices.

\subsubsection{Regularization} We add $L1$-regularization on the shape latent code $\beta \in \mathbb{R}^{236}$ and texture latent code $\theta \in \mathbb{R}^{50}$ to avoid them from being too large. The regularization is formulated as:

\begin{equation}
    L_{reg}=\frac{1}{236}\sum_{i=1}^{236}\left|\beta_{i}\right| + \frac{1}{50}\sum_{i=1}^{50}\left|\theta_{i}\right|
\end{equation}

The overall loss is the weighted sum of the losses above:
\begin{equation}
\begin{split}
L_{total} = \lambda_{contour}L_{contour}+\lambda_{sim}L_{sim}+\lambda_{cam}L_{cam}\\
+\lambda_{lmk} L_{lmk}+\lambda_{photo}L_{photo}+\lambda_{smooth}L_{smooth}+\lambda_{reg}L_{reg}.
\end{split}
\end{equation} where $ \lambda_{contour}=100$, $ \lambda_{sim}=1$, $ \lambda_{cam}=100$, $ \lambda_{lmk}=10$, $ \lambda_{photo}=100$, $\lambda_{smooth}=10$, and $ \lambda_{reg}=0.005$ in our experiments. The regularization is weighted by a small value since large regularization would lead to underfitting of the model. The landmark loss is weighted by 10 instead of 100 since it is less precise than the contour loss. The mesh smoothing could also be regarded as a regularization on meshes so it is also less weighted than other losses.

\section{Appendix B: Training Configurations}

For the depth estimation model, we use $8,000$ synthetic data for training and $2,000$ for validation. We train the model for $100$ epochs using Adam optimizer with an initial learning rate of $0.001$, which is declined by $0.2$ in the $50$th epoch. For the reconstruction model, we randomly split AudioEar2D into training and validation set by $9$:$1$. All left ears are flipped to the right following prior works, so that the model only process one-sided ears. We train AudioEarM for $100$ epochs in two stages. For the first $20$ epochs, we fix the shape and texture latent code so that the model outputs mean shape and texture, and minimize only landmark loss and regularization loss to train the camera and lighting. We empirically find this benefits a reasonable shape reconstruction. For the rest $80$ epochs, we optimize all the parameters by minimizing the weighted sum of all losses. We use Adam optimizer with an initial learning rate of $0.001$ and multiply the learning rate by $0.2$ at the $50$th, $75$th and $90$th epoch.

\section{Appendix C: Impact of Trainset}

\begin{table}[!ht]
    \caption{The impact of trainset. \#2-3 analyze the occlusion of AudioEar2D. \#4-5 compare AudioEar2D and IBug-A. \#6 uses synthetic data to train the reconstruction model with explicit latent code supervision instead of using it in the depth-guided way.}
    \vspace{-3pt}
    \begin{tabular*}{\hsize}{@{}@{\extracolsep{\fill}}c|lcc@{}}
    \toprule
    \# & Trainset & Scale & S2M \\
    \midrule

    1 & AudioEar2D & 2,000 & \textbf{1.28mm}\\
    2 & \#1 $-$ earrings & 1,682 & 1.36mm \\
    3 & \#1 $-$ hair \& earrings & 849 & 1.44mm \\
    4 & IBug-A  & 605 & 1.48mm \\ 
    5 & \#1 $+$ IBug-A  & 2,605 & 1.45mm \\
    6 & \#1 $+$ synthetic & 12,000 & 1.63mm \\
    
    \bottomrule
    \end{tabular*}
    \label{tab:ablation_data}
\end{table}
We examine the impact of trainset by training AudioEarM on different trainsets. First, we remove the partly occluded ears in AudioEar2D from the trainset, to examine the impact of occlusions (Table \ref{tab:ablation_data} \#2-3). We observe that as we remove more occluded images, the performance gets worse. It seems that partial occlusion does not affect the performance as much as a smaller trainset does. Then we compare AudioEar2D with IBug-A (\citeauthor{zhou2017deformable}), a prior ear dataset with landmarks. IBug-A contains $605$ images compared to $2,000$ in AudioEar2D (Table \ref{tab:ablation_data} \#4-5). Training AudioEarM only on IBug-A yields a result of $1.48$ mm ($1.28$ mm on AudioEar2D). It demonstrates that a smaller scale of trainset does lead to performance degradation, which proves the value of AudioEar2D. We also combine AudioEar2D with IBug-A together for cross-domain training and find that the performance is worse than training on AudioEar2D alone, although the trainset scale is larger. We reckon it is caused by a domain gap between the two datasets. Specifiaclly, the \textbf{domain gap} between IBug-A and our AudioEar2D could be in two aspects. One is that IBug-A includes grayscale images, while AudioEar2D is entirely RGB. The other is about the landmark annotation standard. Since some of the landmarks lack explicit definition, our annotation standard might be different from theirs, which might leads to a domain gap. We reckon the first factor is more critical to the degraded performance since our contour loss could alleviate the second one. Lastly, instead of using the synthetic data to train a depth estimation model, we combine it with AudioEar2D together as the trainset and use the known latent code of the synthetic data as explicit supervision for the training of reconstruction model. This practice is similar to the MediaPipe Hands (\citeauthor{zhang2020mediapipe}). It yields a result of $1.63$ mm, which is rather poor. In this respect, the depth estimation model in our method might be playing the role to bridge the gap between the synthetic and realistic data. 

\section{Appendix D: SPL Error}
\begin{figure}[h!]
    \centering
    \includegraphics[width=\linewidth]{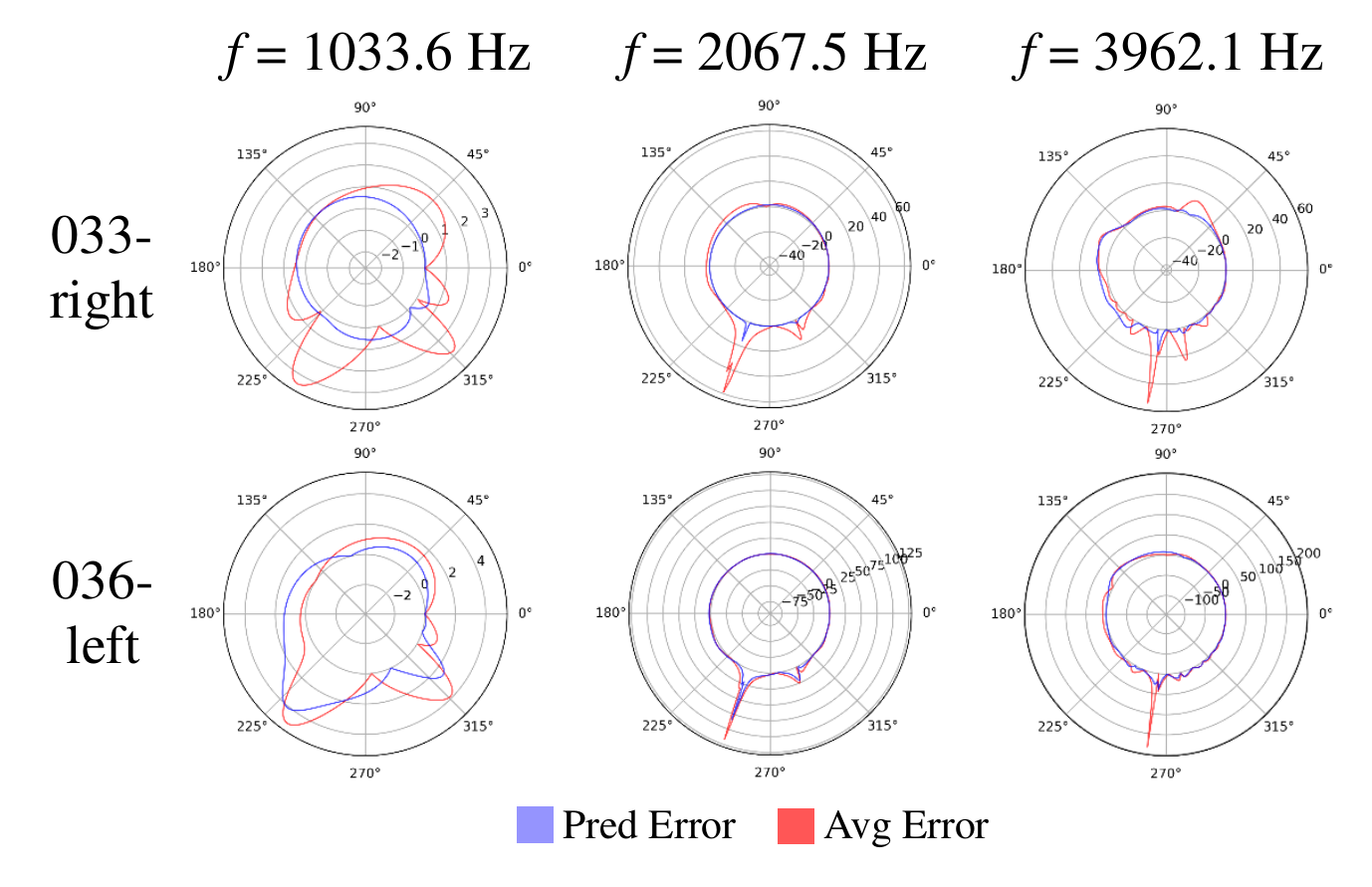}
    \caption{Illustrations of the SPL error in polar coordinates under three simulation frequencies. Polar angles denotes the simulated azimuths, while radius denotes the magnitude of error. The mean SPL error is calculated among all azumith angles.}
    \label{fig:spl_error}
\end{figure}

For a better understanding on the SPL error, we illustrate it with polar coordinates in Figure \ref{fig:spl_error}. HRTF is simulated in all azimuth angles ($361$ angles) on the horizontal plane. Then the SPL error is drawn in a polar coordinate system with  azimuth angles as angular coordinates and the error in each angle as radial coordinate. The final SPL error is averaged among all angles.

\section{Appendix E: Limitations}

Although the reconstruction from single-view images in this work has obtained a relatively good result compared to the average ear, there is still room for further improvements in order to obtain a better spatial audio effect. In future work, we will consider a multi-view reconstruction setting. More views could reveal more 3D structures of the ears and lead to better reconstruction.

The HRTF is not only conditional on the ear structure, but also on the head and body structure. Besides, the position and orientation of the reconstructed ears with respect to the head are not automatically estimated. Therefore the simulated HRTF is not the real HRTF of the individual. In future work, we are going to jointly reconstruct the human body and estimate the relative position of ears at the same time to obtain a more realistic HRTF.

\end{appendices}
\end{document}